\documentclass[preprints,article,accept,moreauthors]{Definitions/mdpi}

\firstpage{1} 
\pubvolume{1}
\issuenum{1}
\articlenumber{0}
\pubyear{2026}
\copyrightyear{2026}
\datereceived{ } 
\daterevised{ } 
\dateaccepted{ } 
\datepublished{ } 

\Title{A Dataset-Intrinsic X-Factor Affecting Model Performance}

\Author{Josiah Couch \textsuperscript{1}\orcidA{},  Miao Li \textsuperscript{1}\orcidB{}, Rima Arnaout \textsuperscript{2}\orcidC{} and Ramy Arnaout \textsuperscript{1,}*\orcidD{}}

\AuthorNames{Firstname Lastname, Firstname Lastname and Firstname Lastname}

\address{%
\textsuperscript{1} \quad Department of Pathology, Beth Israel Deaconess Medical Center, 330 Brookline Ave., Boston, MA 02215, USA\\
\textsuperscript{2} \quad Department of Medicine, University of California San Francisco, San Francisco, CA 94143, USA}

\corres{Correspondence: \texttt{rarnaout@bidmc.harvard.edu};}

\abstract{
In the quest to optimize the performance of machine-learning classifiers, three factors---model architecture, dataset size, and class balance---have been shown to influence test-time performance but do not fully account for it. Previously, evidence was presented for an additional factor accounting for the residual, but it was unclear whether this residual was actually a joint property of the dataset and the model architecture, or an intrinsic property of the dataset itself. If the residual is truly dataset-intrinsic and independent of model architecture, dataset size, and class balance, then the same datasets should perform better (or worse) regardless of these other factors. To test this hypothesis, here we create thousands of datasets, each controlled for size and class balance, and use them to train classifiers with a wide range of architectures, from random forests and support-vector machines to deep networks. We find that classifier performance correlates strongly by subset across architectures ($R^2=0.82$), supporting this residual factor as an intrinsic property of datasets independent of dataset size and class balance and of model architecture. Our results suggest that this dataset-level residual is also apparent in datasets' constituent classes. This residual ``X-factor'' joins size, class balance, and model architecture as an independent correlate of performance and a separate target for optimizing machine-learning-based classification.}


\begin{document}

\section{Introduction}\label{sec: introduction}

Since the initial observation that neural-network models obey certain scaling laws \citep{hestness2017}, there have been many empirical \citep{Ahmad1988,prato2021,althnian2021,caballero2023,gpt4} and theoretical \citep{Seung1992, bahri2024, Maloney:2022cvb, michaud2024, su2024, Zhang:2024mcu, bordelon2024, atanasov2024, levi2024, brill2024} studies examining the effects of training-set size and class balance (measured as the entropy of the class frequencies), model complexity (the number of model parameters), and training effort (amount of compute) on performance (accuracy or error). However, these factors do not fully account for the variance in performance observed among models trained on different training sets. It stands to reason that additional factor(s) could be involved.

Consider the well-known (but possibly apocryphal) anecdote about an image classifier trained to look for tanks \citep{riley2019}. The story goes that while the model achieved 100\% test accuracy during training, it failed on real-world examples, because in the training set, all the images with tanks also contained clouds, and all the images without tanks lacked clouds. It is no surprise that a classifier trained on such a training set would perform poorly on real-world examples, regardless of the size or class balance of the training set, the complexity of the model, or the duration of training. However, relatively few studies have sought to define this residual factor. To this end, a reasonable first question is: is this ``X-factor'' intrinsic to the training set and thus universal across models, or do different models require training datasets with different characteristics to fully exploit their potential? The answer determines whether to devote more attention to the dataset or to the model. Here, we address this question to better understand this X-factor: that part of classifier performance that is not explained by the dataset's size or class balance or other aforementioned factors.

Specifically, we test the null hypothesis that once size and class balance are controlled, there will be no correlation between the performance of unrelated models across different training sets beyond what would be expected by chance. We present results that reject this hypothesis, and thereby demonstrate that this X-factor is dataset intrinsic. Although we speculate about causal feature(s) or characteristic(s), further investigation of this X-factor is left for future work.

In the context of prior work, this X-factor is most properly linked to the general concept of dataset quality. However, ``quality'' is a general term that has been used to refer to everything from labeling accuracy to data completeness to image resolution, and is also tied to dataset size and class balance. \citet{rahal2025enhancingmachinelearningperformance} described ``quality'' as any residual contributor to model performance once these other factors have been accounted for---what we refer to here as the X-factor, as its nature is unknown. A 2024 study \citep{couch2024} proposed that the X-factor is an intrinsic property of training sets, and specifically that it could be measured using a sentropic \citep{nguyen2026} quantity called the metacommunity alpha diversity, part of the recently introduced similarity-aware-diversity framework \citep{leinster2012, reeve2016, leinster2021} referred to in that work as LCR. This diversity is interpreted as the effective number \citep{leinster2021} of image-class (or image-label) pairs in the training set, after accounting for the pairwise similarities and differences among images. (The presence of similarity means that the \textit{effective} number is never greater than the \textit{actual} number; similarity decreases diversity.) 

Diversity and (s)entropy are related by exponentiation \citep{leinster2021}. That an information-theoretic measure might be related to training-set quality is reasonable in the context of the tanks-and-clouds example. That dataset has two separate ``labels:'' one that is being learned---tank vs. no tank---and a second that is irrelevant to the classification task---clouds vs. no clouds. If the training-set images were labeled for both tanks and clouds, the maximum-possible Shannon entropy of their joint distribution would be 2 bits---a diversity of $2^2=4$ classes. This maximum would be achieved by balancing a training set across all four combinations: tank/cloud, no-tank/cloud, tank/no-cloud, and no-tank/no-cloud. However, the entropy of the actual training set in the anecdote is only 1 bit (2 classes)---tank/cloud vs. no-tank/no-cloud---due to the (unintentionally) perfect correlation between these labels. This example suggests that it is reasonable to think that the quality of the training set should be related to its entropy/diversity.

While that study demonstrated a correlation between model performance and metacommunity alpha diversity, it only evaluated a single model architecture, ResNet-18 \citep{couch2024}. A more systematic test of whether quality is an intrinsic property of training sets requires comparing performance across multiple models. In addition, the subsets in the 2024 study were all constructed via random sampling (conditioned on a class balance which was separately sampled from a complicated distribution). However, it is fair to expect that, conditioned on the class balance, the performance will be similar for any subset created by uniformly random sampling---i.e., that all uniformly random subsets will lead to similar model accuracy, and that their accuracies will cluster tightly around the mean value---leaving not enough variance in performance to test the hypothesis.

To overcome this limitation, our approach is to select subsets in a way that biases model performance away from the mean. Prior work has described several methods for subset selection, including active learning \citep{houlsby:2011, gal:2017, ren:2021, kadota:2022, kirsch:2022a, kirsch:2022b, mindermann:2022, jesson:2022, huang:2022, riis:2022, ziatdinov:2022, khosravani:2022}, instance selection \citep{Chinn2021, chinn2023a, Xie:2024}, dataset pruning \citep{Sorscher:2022, Maharana:2023}, and coreset selection \citep{har-peled:2007, killamsetty:2021, guo:2022, griffin:2024, chen:2025, zheng:2025}. For example, a 2022 study \citep{Sorscher:2022} showed that the performance obtained by training on a randomly sampled subset can be improved upon by choosing either the ``easiest'' (for small datasets) or ``hardest'' (for large datasets) examples from a larger parent dataset. However, in this study, easiest and hardest are defined with respect to a ``probe'' (i.e., a weakly trained) perceptron classifier, which had to be trained prior to selection.

Here, we employ a subset-selection approach that eliminates the need for pretraining by selecting images based on their root mean square differences (RMSDs). This approach is inspired by sentropy, in which a dataset's information content is a function of the pairwise similarities among its elements. Using this approach, we find that the performance of subsets indeed correlates very strongly (though not perfectly) across model architectures, falsifying our null hypothesis and thereby supporting the conclusion that the X-factor of residual quality unexplained by size or class balance (which we control) is intrinsic to datasets, not the model or dataset-model pairs.

\section{Materials and Methods}\label{sec: methods}

\begin{figure*}[!ht]
    \centering
    \includegraphics[width=0.8\linewidth]{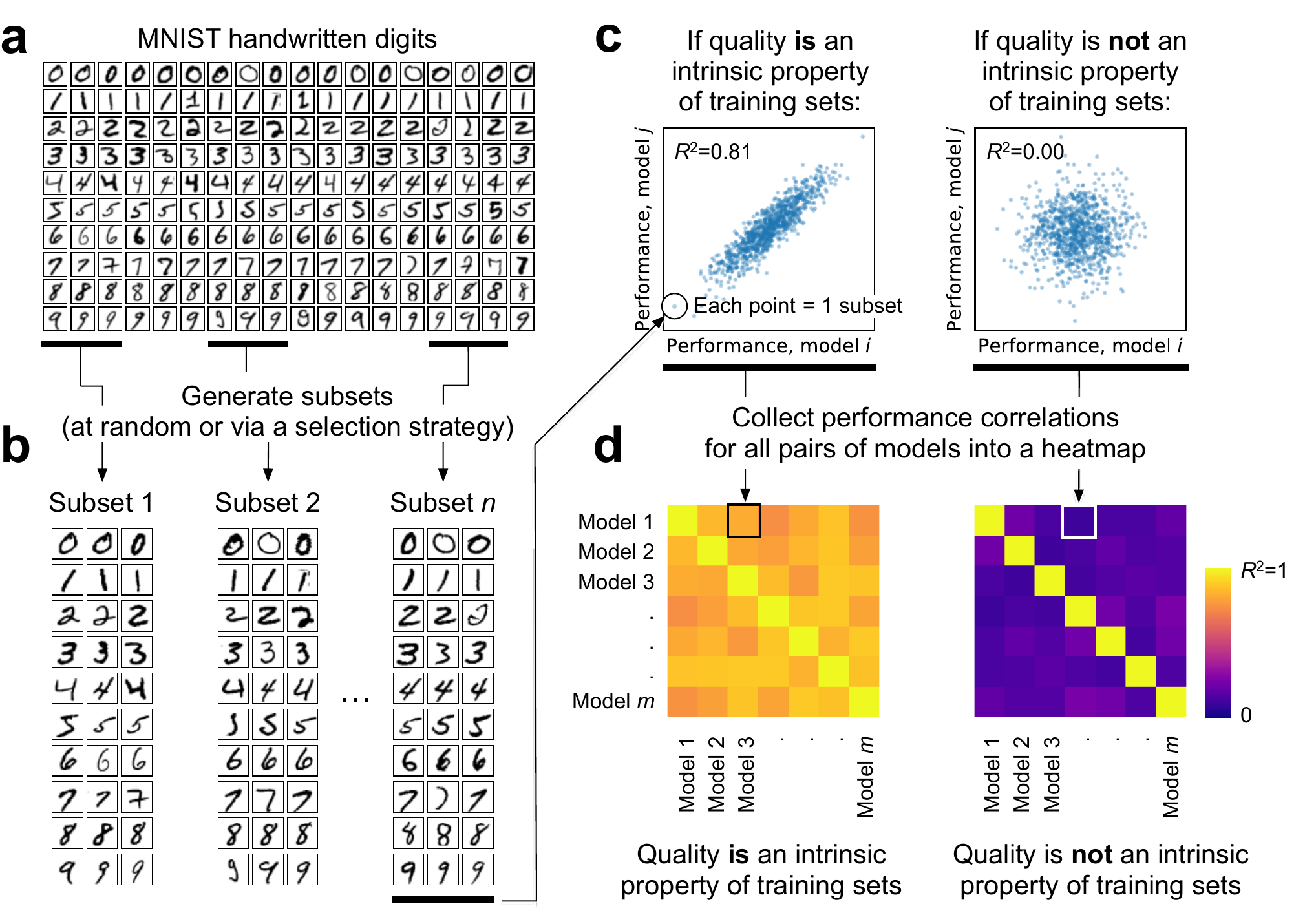}
    \caption{Schematic overview of research strategy. (a) Beginning with MNIST, (b) generate many subsets, either at random (``random subsets'') or according to one of several selection strategies (``selected subsets''), (c) compare how different models score these subsets, and (d) collect all correlations (e.g. into a heatmap). If performance is strongly correlated regardless of model---i.e. if the same subsets lead to the same relative performance for all model pairs---this provides direct evidence against our null hypothesis, and thus evidence for the conclusion that there is an X-factor intrinsic to training sets.}
    \label{fig: overview}
\end{figure*}

\subsection{Strategy (Fig. \ref{fig: overview})}
For each dataset, we first generated random subsets of various sizes uniformly at random, but conditioned on their being perfectly class balanced, and characterized the performance of different models (see below) when trained on these subsets (Fig. \ref{fig: overview}a). Next, we generated additional class-balanced subsets of the same size, such that models trained on these subsets differed in performance from the uniformly random subsets (Fig. \ref{fig: overview}b). Finally, we tested whether differences in performance correlated across models (Fig. \ref{fig: overview}c-d), which directly provides evidence about our null hypothesis. 

\subsection{Models}

To support the generality of our results, we tested diverse model architectures: four classical architectures---random forest (RF), support-vector classifier (SVC), Gaussian naive Bayes (GNB), and K-nearest neighbors (KNN)---and three neural-network architectures---multilayer perceptrons (MLP), ResNets (ResNet-18 and ResNet-34 \citep{resnet18}), and a vision transformer (ViT) \citep{dosovitskiy_image_2021}. Scikit-learn \citep{scikit-learn} was used for the classical models and PyTorch \citep{pytorch} for the neural nets. The Appendix provides details on how each model was run.

\subsection{Datasets}

We consider five ``parent'' datasets: a relatively simple benchmark, MNIST \citep{mnist}, and four much more complex real-world medical datasets: BloodMNIST, PathMNIST, OrganAMNIST, and OrganCMNIST from the MedMNIST \citep{medmnistv1, medmnistv2} collection of medical imaging datasets.

All the training sets in this work are subsets of the preselected training sets of one of these five parent datasets. For a given parent dataset with $k$ class labels and a fixed size of $n$=10, 30, 100, 300, or 1,000 images, we generated one group of subsets of size $k\times n$ by choosing $n$ images of each class uniformly at random. We refer to this baseline group as ``random subsets.'' We generated other groups (also of size $k\times n$) by oversampling $n+m$ images from each class and then selectively discarding the $m$ excess images according to one of several selection strategies (below). We refer to these as ``selected subsets.''

For each parent dataset, we generated subsets according to five strategies: ``random,'' ``uniqueness,'' ``global mean,'' ``other mean,'' and ``own mean'' (see Appendix \ref{app: selection}). 100 random subsets were first generated for each of our five sizes $n$. For the other four strategies, five subsets were generated for each of the 10 choices of selection strength $m$ (for $m=$ 6, 10, 20, 30, 60, 100, 200, 300, 600, or 1,000) and selection direction (positive vs. negative), leading to up to $4\times5\times10\times2=400$ selected subsets at each size for each parent dataset. Thus, in the case of MNIST, we generated $100+400=500$ subsets at each of five unique sizes, resulting in a total of 2,500 subsets. For some MedMNIST datasets, the number of images available in the smallest class limited the achievable sizes and selection strengths, leading to a reduced set.

All selection strategies relied on the pixel-wise root-mean-square-difference (RMSD) distance between images. Three of the strategies calculated and made use of the mean training image per class, which for each class $k$ was the $28\times28$-pixel image that is the pixelwise mean of all the images in the relevant training set that belong to class $k$ (in the case of MNIST, the mean over all 6,000 images representing a given digit). Note that the content of the test set did not influence these selection strategies.

\subsection{Training and testing}

The training of the sklearn models (three variants of KNN, GNB, SVC, and RF) was performed using the default settings of \verb|model.fit(X,y)|. Training of neural networks (ViT, ResNet-18, ResNet-34, and MLP-1 through 5) was done using cross-entropy loss and the Adam optimizer \citep{kingma2017} with a batch size of 200 (note that for the smallest subsets, which had only 10 images per class, this results in only a partial batch per epoch). All models except for ViT were trained for at least 50 epochs to a training set accuracy of 99\%, or to a maximum of 500 epochs, whichever came first. As ViTs tend to be data-hungry and struggle with smaller datasets \citep{pandey_are_2023}, the standard was relaxed to a maximum of 50 epochs or achieving an accuracy of 99\% for the training to finish in a reasonable time frame. The learning rate was initially set to 1e-3, was reset to 1e-4 on the second epoch, and was further reset to 1e-5 if the 300th epoch was reached. All models trained on these subsets were tested against the predefined test set associated with the relevant parent dataset. Performance was assessed by the accuracy and the error (= $1-\text{accuracy}$). Note that because all training sets were class balanced, and the MNIST test set was also class balanced, accuracy and balanced accuracy are the same for the MNIST models. This is not necessarily the case for the MedMNIST models.

\subsection{\textit{Z}-scores}

We computed the mean and standard deviation of the error for training sets by subset size and for each model. We did this first for random subsets, as a baseline. We computed a \textit{Z}-score for the error on each random subset by subtracting the relevant mean error and dividing by the relevant standard deviation. To compare selected subsets to random subsets, we then computed a \textit{Z}-score for the error on each selected subset using the mean and standard deviation from the relevant random subsets (by size and model).

\subsection{Correlations and confidence intervals}

We computed $R^2$ (the square of Pearson's $R$) of \textit{Z}-scores for different models and (groups of) subsets. We bootstrapped the relevant set of subsets with 1,000 resamples to obtain 95\% CIs on these $R^2$s.

\section{Results}\label{sec: results}

\begin{figure}[ht]
    \centering
    \includegraphics[width=1.\linewidth]{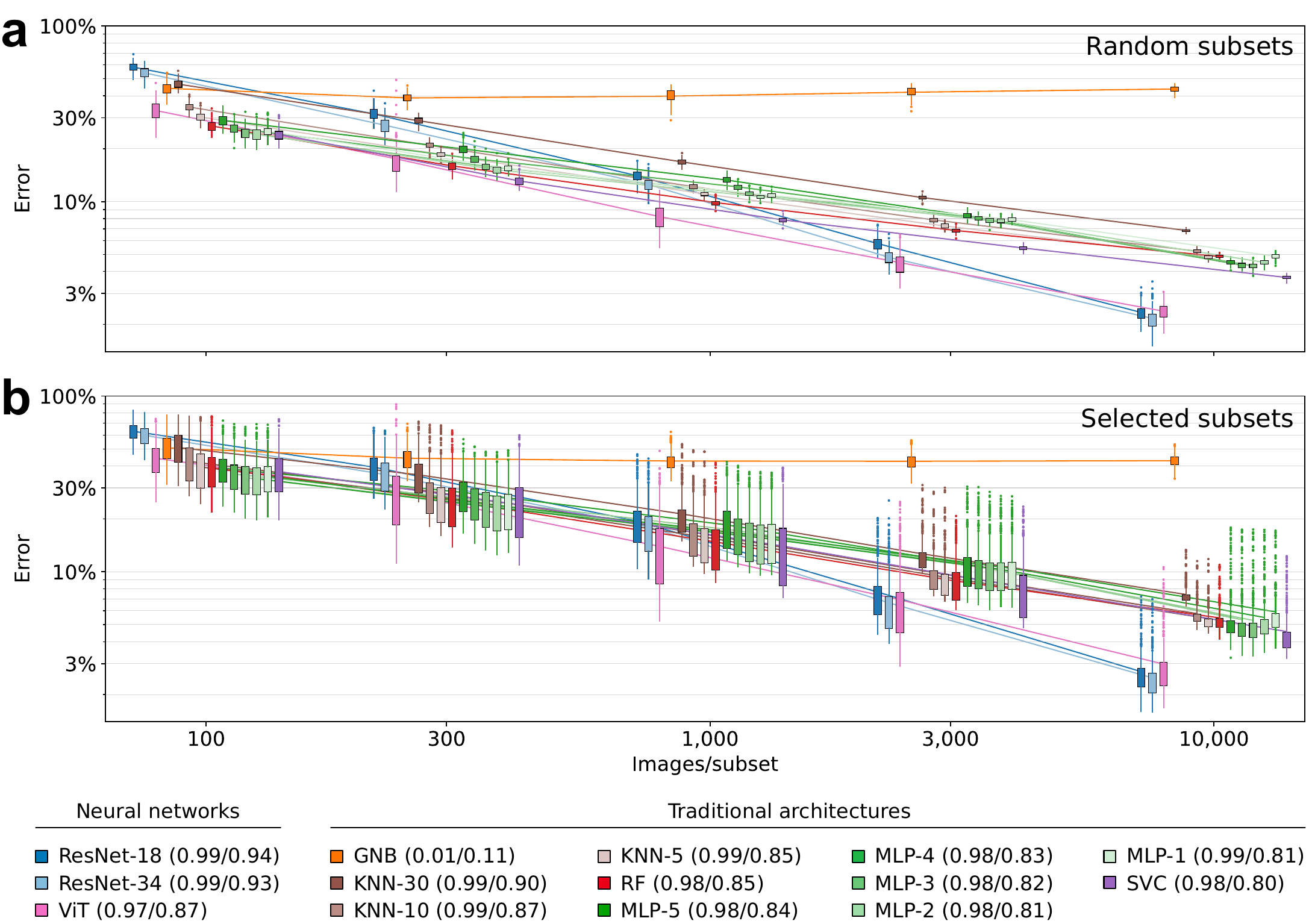}
    \caption{Performance of 14 models on all (a) random and (b) selected subsets vs. subset size. Models include two different ResNets (blue), five different MLPs (green), and three different KNNs (brown). At each size, a box-and-whiskers element shows the range of errors, with any outlier points plotted separately. For clarity, elements are jittered along the x-axis; thus, each horizontal grouping of symbols reflects same-sized subsets. Whiskers are wider for selected subsets (b) than random subsets (a), meaning random subsets' performance varied less than selected subsets' performance. Line segments connect the geometric means at each subset size. The relatively straight lines on these log-log plots demonstrate known scaling laws of performance with training-set size. Note the slopes are steeper for more expressive models: traditional models have lower error than neural-network models at small subset sizes, but neural-network models outperform traditional models at large sizes. Legend: numbers in parentheses are the $R^2$ for power-law fits for each model in panels (a) / (b).}
    \label{fig: performance}
\end{figure}

We find strong evidence against our null hypothesis across highly heterogeneous models and five ML datasets. In particular, we trained seven different types of models and 14 total variants, on each of 2,500 MNIST subsets (Fig. \ref{fig: overview}) and a smaller number of subsets from each of PathMNIST, BloodMNIST, OrganAMNIST, and OrganCMNIST (limited only by the sizes of these datasets). Four models (six variants) were classical (KNN for $k\in\{5,10,30\}$, GNB, RF, and SVC) and three models (eight variants) were neural networks (ViT, five MLPs, and two ResNets; see Methods). The 2,500 MNIST subsets consisted of 500 subsets at each of five different sizes: 100, 300, 1,000, 3,000, and 10,000 images per subset, corresponding to 10, 30, 100, 300, and 1,000 images per class for each of the 10 classes (Fig. \ref{fig: overview}a-b). Critically, all subsets were perfectly class balanced, with the same number of images for each class, eliminating size and class balance as possible contributors to model performance. At each size, 100 subsets were generated by sampling images in each class uniformly at random (see Methods), and 400 were selected according to one of several selection strategies to introduce different kinds of variation while maintaining maximal class balance.

Results for MNIST were representative of results for the MedMNIST subsets. Therefore, for brevity, we describe the former in detail in the main text, and present details for the latter in Appendix (\ref{sec: medmnist_res}).

Performance on random subsets varied by model and by subset size, recapitulating familiar log-log scaling laws with size, with steeper slopes for the more expressive models (Fig. \ref{fig: performance}a). The only exception to this pattern was GNB, which showed negligible trend with size, likely owing to its being too simple a model even for MNIST classification. Overall, error rates ranged widely, from 1.5\% to 70\%, by model and size. The classical models RF and SVC were the best performers on the smallest subsets, while the ResNets, followed closely by ViT, performed best on the largest. For subsets of a given size trained on a given model, performance clustered very tightly around the mean for these random subsets: model and subset size generally explained more than 98\% of the performance, leaving less than 2\% residual variance.

\begin{figure}
    \centering
    \includegraphics[width=0.9\linewidth]{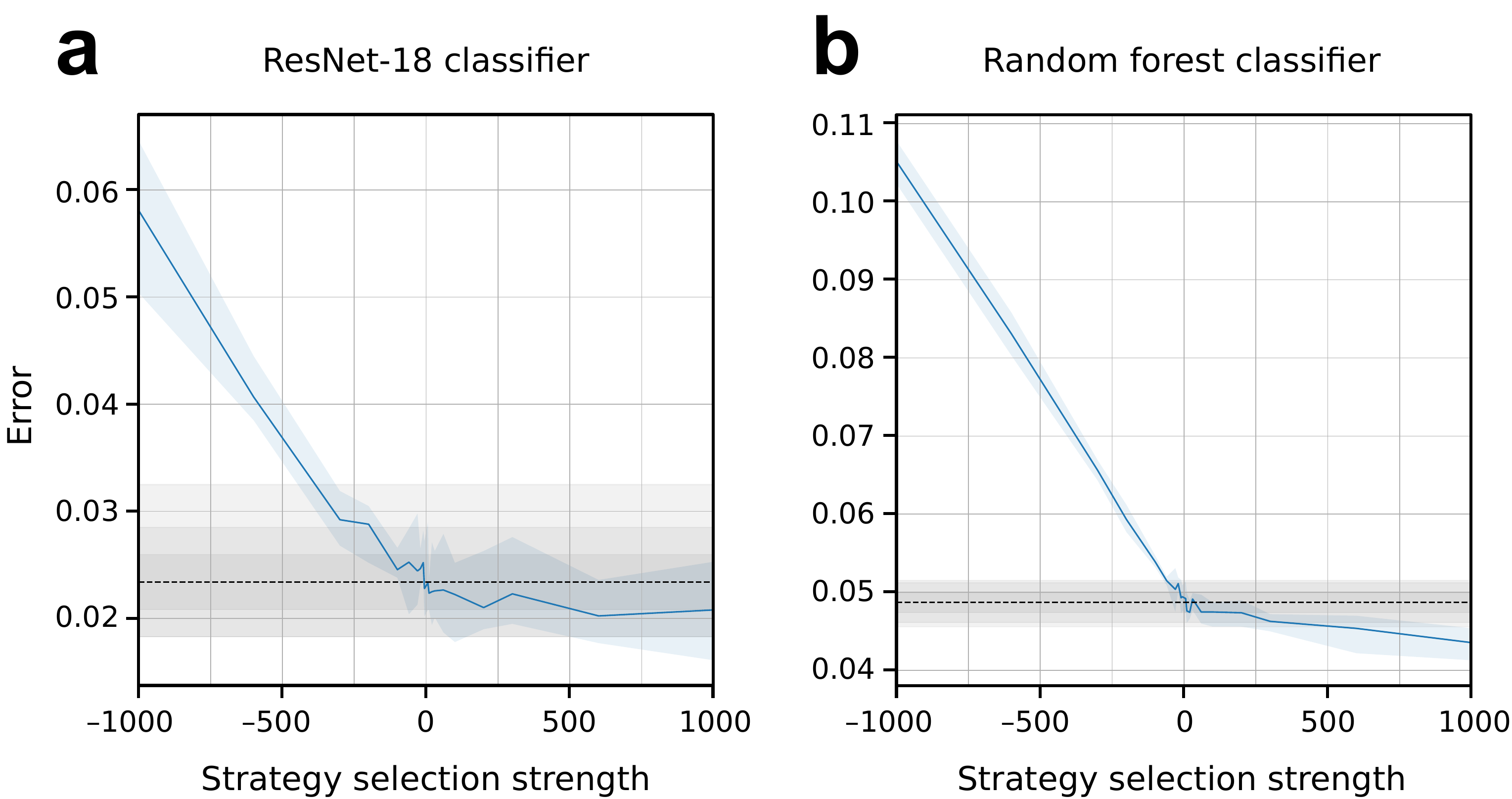}
    \caption{Error of (a) ResNet-18 classifiers and (b) random forest classifiers trained on 10,000-image subsets selected through the uniqueness strategy, as a function of the oversampling/selection strength. Note that negative values indicate negative selection. Blue lines indicate the means of the performance of selected subsets at the selection strength indicated on the x-axis; blue contours indicate the range of performance for these subsets. Black dotted line is the mean error for random subsets; the dark-to-light gray contours correspond to $\pm$1 s.d., $\pm$2 s.d.s, and entire range, respectively, for the random subsets. Note the blue lines intersect the black dotted lines at zero because zero selection strength is the same as random selection.}
    \label{fig: err vs selection}
\end{figure}

In contrast to the random subsets, performance on the selected subsets ranged more widely for each subset size and for each model (Fig. \ref{fig: performance}b). Model and subset size together explained only 80-94\% of the variance in performance on these subsets ($R^2$ from fits in Fig. \ref{fig: performance}b), leaving 6-20\% residual variance unexplained (again, excepting GNB). Note that the goal of selected strategies was to expand the range of performance to make correlations across models easier to detect, not to achieve better performance than that obtained by the best random subsets, and indeed they were usually worse. Therefore, it was interesting that selected subsets did occasionally outperform, suggesting fortuitous avenues for future dataset optimization (Fig. \ref{fig: err vs selection}).

Specifically, the selection strategy that outperforms random sampling most often and by the widest margin is the ``uniqueness'' strategy. A detailed description of this strategy (and the others) can be found in the supplemental materials in the technical appendix (Appendix \ref{app: selection}). Roughly, this strategy is designed to select highly unique (positive) or highly redundant (negative) subsets of images. Fig. \ref{fig: err vs selection} shows that for positively selected subsets, uniqueness improves the average performance with increasing strength, across models. Consistent with this finding, negative selection reverses this effect, leading to worse-than-random performance.

\begin{figure}[t]
    \centering
    \includegraphics[width=1.\linewidth]{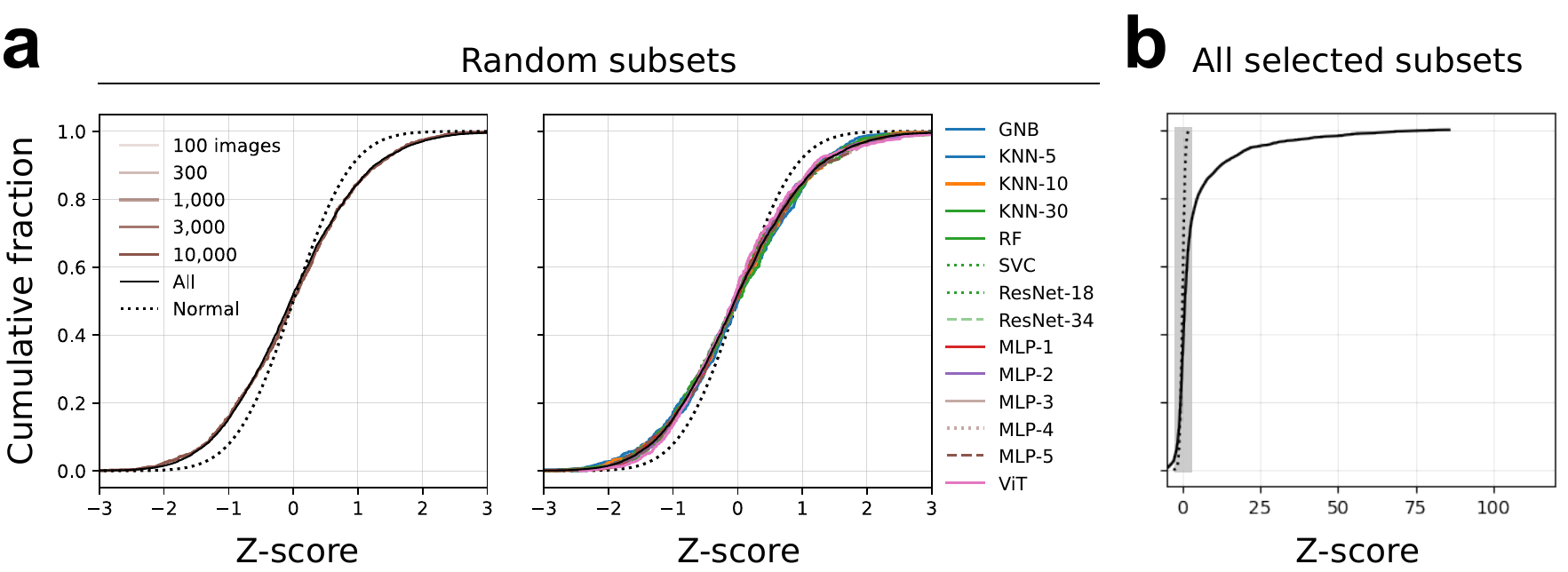}
    \caption{Scaled residual variance in performance from Fig. \ref{fig: performance}, plotted as cumulative density functions (CDFs) of \textit{Z}-scores, which measure how much a subset's error deviates from the expected error of all the subsets of the same size for the same model, normalized by variance. (a) CDFs for random subsets pooled by subset size (left) and model (right) plotted against the CDF for a normal (Gaussian) distribution (dotted black line). Note their uniformity regardless of model or subset size, and how they differ subtly but reproducibly from the normal CDF (Kolmogorov-Smirnov $p<10^{-5}$). Colors and linestyles are the same as in Fig. \ref{fig: performance}. (b) CDF for all selected subsets (solid black line) plotted against the CDF for a normal distribution (dotted black line). Note the much larger range of \textit{Z}-scores for the selected subsets; the x-axis range for each of the plots in (a) is shown in gray in (b) for comparison.}
    \label{fig: distr}
\end{figure}

The performance of random subsets clustered so tightly by size and model that we decided to examine them more closely. To compare these tight distributions to each other, we converted them from errors to \textit{Z}-scores by subtracting off the mean error for each distribution---i.e., for each model and subset size---and then dividing the result by the standard deviation of that distribution. Surprisingly, we found the resulting distributions of residuals were indistinguishable from each other, regardless of model, even for very different model architectures (e.g., RF vs. ViT): they overlapped each other almost perfectly regardless of size or model (Fig. \ref{fig: distr}a). We expected these residuals might follow a normal (Gaussian) distribution; however, we found instead that they differed significantly from normal (Kolmogorov-Smirnov $p<10^{-6}$), specifically in being slightly but persistently flattened. In contrast, the residual variance for the selected subsets was highly skewed (Fig. \ref{fig: distr}b). This pattern supports the conclusion that random subsets are in some fundamental sense all alike, but that outlier subsets with very different performances can be found using appropriate selection strategies.

\begin{figure}
    \centering
    \includegraphics[width=1.\linewidth]{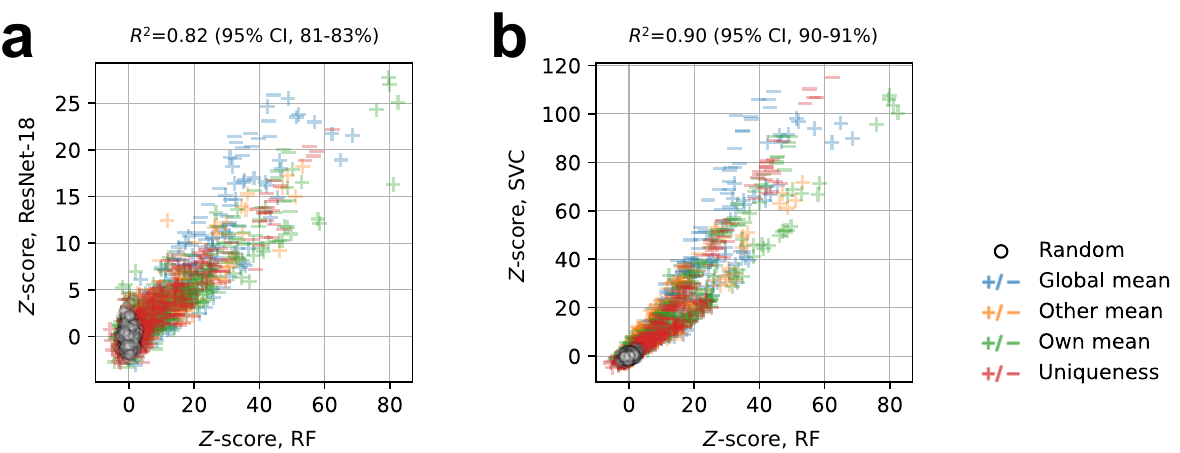}
    \caption{Training sets that perform better/worse by RF also tend to perform better/worse by (a) ResNet-18 and (b) SVC, two representative examples from among the 45 model pairs we compared. Colored + and - symbols, positively and negatively selected subsets, respectively. Note the random subsets cluster near \textit{Z}-scores of zero, with only a few selected subsets outperforming them (corresponding to \textit{Z}-scores below and to the left of them).}
    \label{fig: pairwise}
\end{figure}

\begin{figure}[ht]
    \centering
    \includegraphics[width=1.\linewidth]{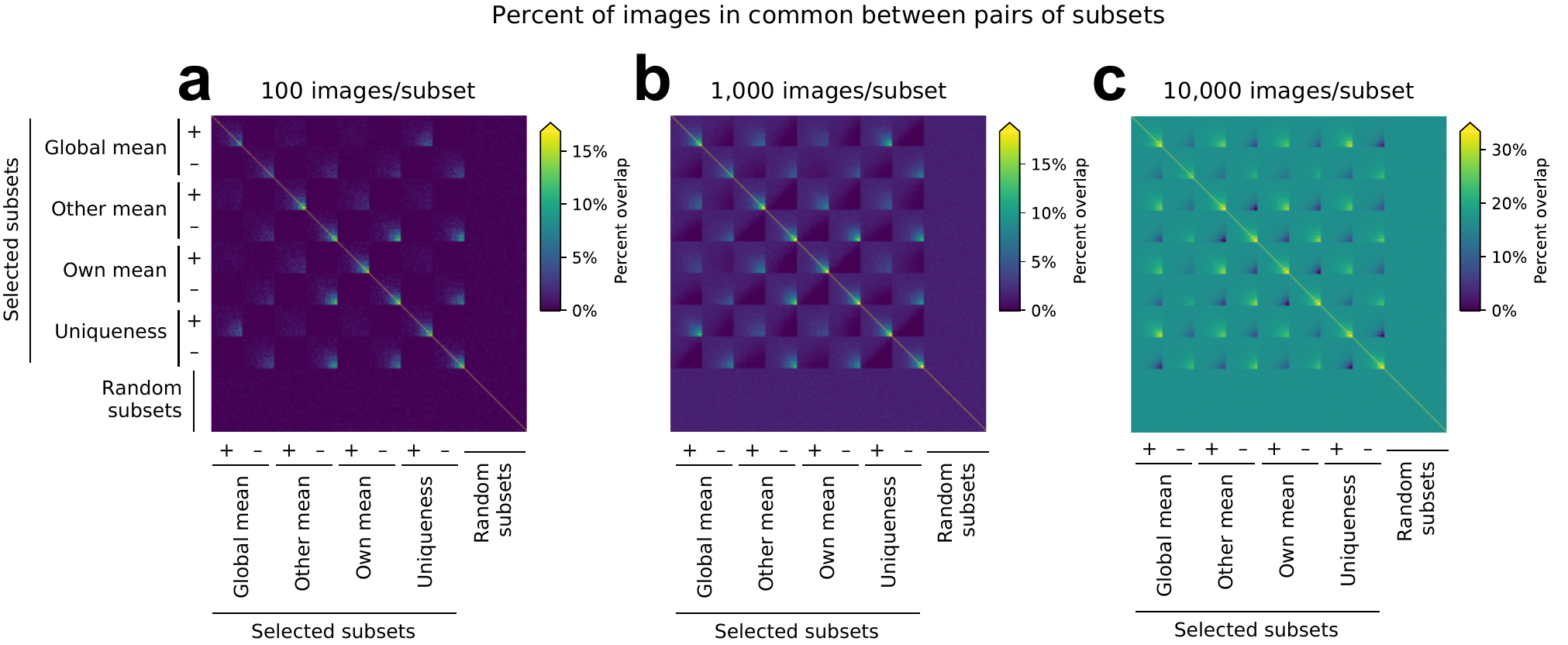}
    \caption{Percent of images that pairs of subsets have in common, for (a) 100, (b) 1,000, and (c) 10,000-image subsets. Positively selected subsets share images more than other pairs, yielding the faint checkerboard pattern. The color scale is saturated at the upper end to better visualize these effects.}
    \label{fig: overlap}
\end{figure}

Having demonstrated that selected subsets exhibit a wide range of performance, as measured by error or \textit{Z}-score, we next tested whether the different models performed better (or worse) on the same subsets. We did this by calculating Pearson's correlation coefficient on the \textit{Z}-scores for subsets according to each pair of models (Fig. \ref{fig: pairwise}). The resulting $R^2$ values (computed as the square of the Pearson's correlation) were high, even between models of radically different architectures such as ResNets vs. traditional models such as RF and SVC. This was true regardless of subset size and selection strategy. The $R^2$s were not due to selected subsets accidentally sharing a high percentage of images, as percent overlap between pairs of subsets was generally very low (Fig. \ref{fig: overlap}).

\begin{figure}
    \centering
    \includegraphics[width=0.95\linewidth]{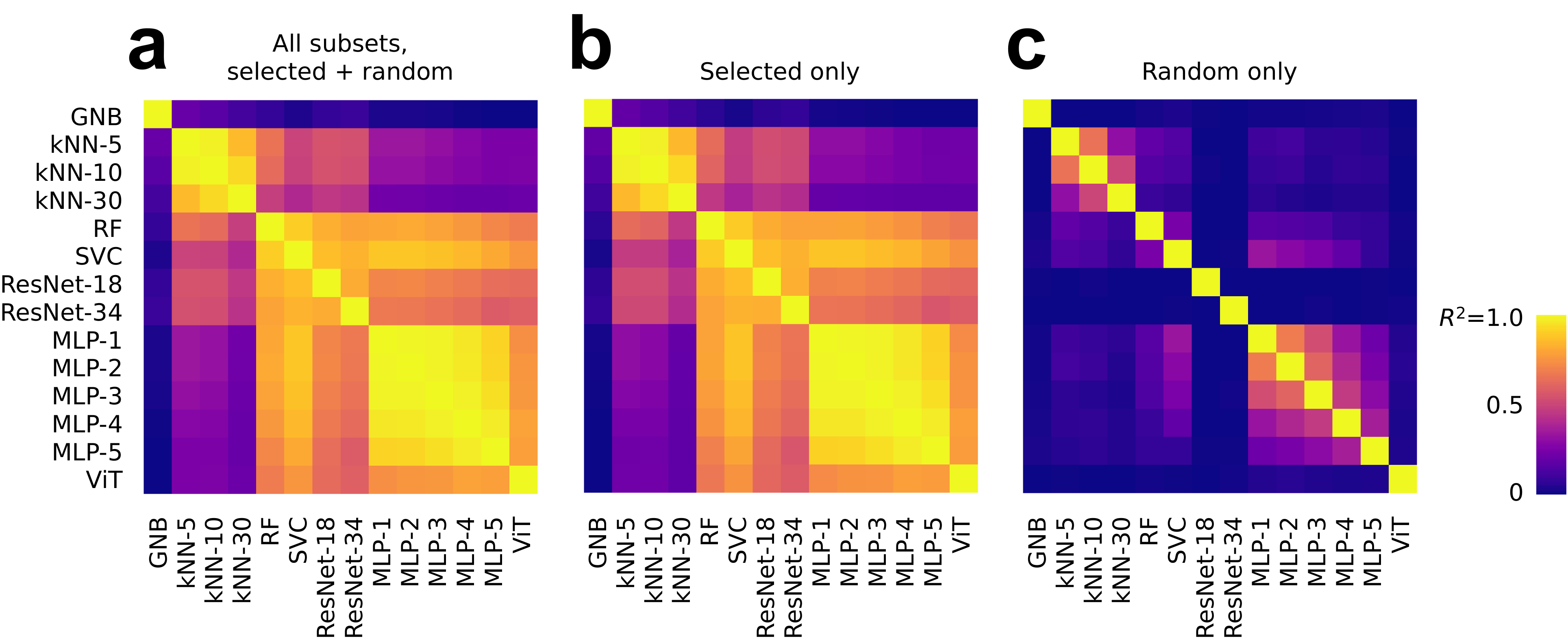}
    \caption{Performance correlations for all pairs of models for (a) all subsets, (b) selected subsets only, and (c) random subsets only. $R^2$ from comparisons such as those in Fig. \ref{fig: pairwise} are shown. The pattern in (a) is driven by the selected subsets, explaining the similarity between (a) and (b). Note this pattern is largely absent from (c), consistent with our demonstration that random subsets exhibit very little variation in performance---too little to detect correlations across models---motivating the selection strategies that generated the selected subsets.}
    \label{fig: subsets}
\end{figure}

Fig. \ref{fig: subsets} collects the $R^2$ values for all pairs of models, illustrating the near-universality of this finding, supporting an X-factor intrinsic to datasets, irrespective of model, falsifying our null hypothesis. Again, the only exception was GNB, which showed low $R^2$ with all other models. Given that GNB's performance was always poor, with error rates of 32-83\% (Fig. \ref{fig: performance}), and improved little if at all even as training-set size increased 100-fold, the likely explanation is that this model's expressiveness is too limited to draw meaningful conclusions; it is the exception that proves the rule. Excluding it, we found the mean $R^2$ over all pairwise model comparisons was 0.82 (95\% CI, 0.80-0.84). Thus, almost all of the residual variance in model performance was indeed explained by a property intrinsic to datasets.

\begin{figure}[ht]
    \centering
    \includegraphics[width=1.\linewidth]{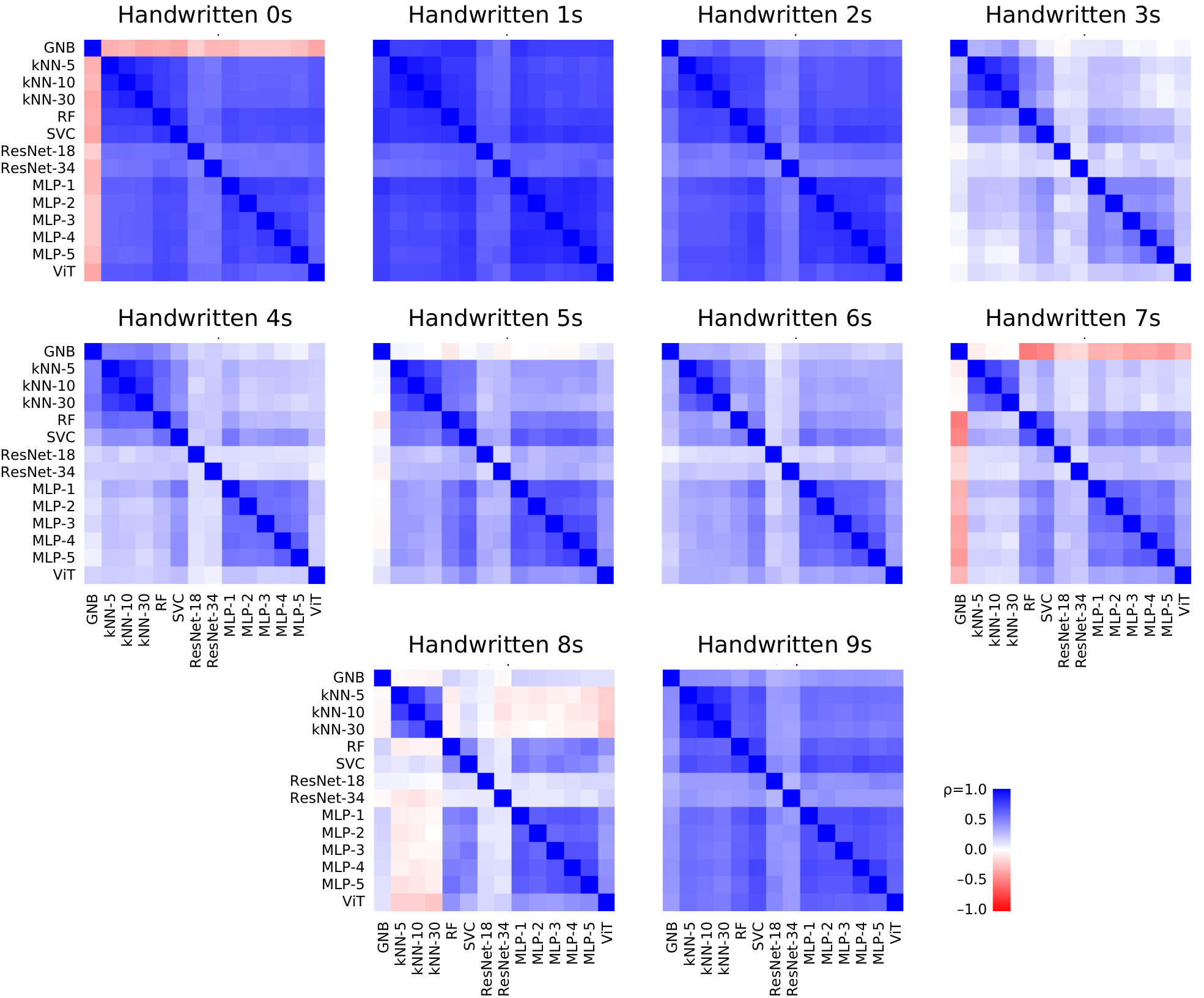}
    \caption{Spearman's $\rho$ for pairwise performance comparisons for each of the 10 classes in MNIST, across all subsets.}
    \label{fig: classes}
\end{figure}

\begin{figure}[ht]
    \centering
    \includegraphics[width=0.75\linewidth]{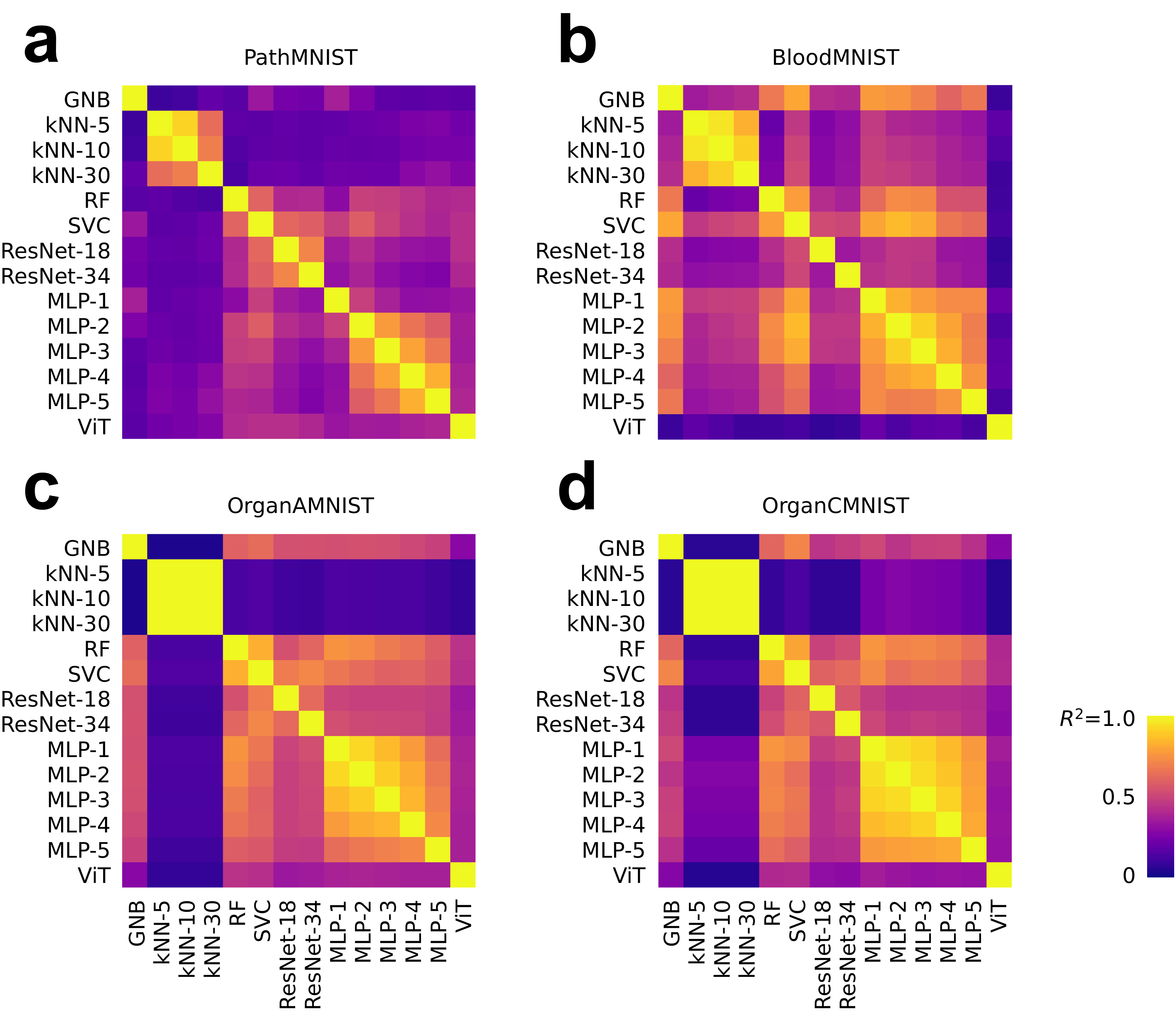}
    \caption{Performance correlations for all pairs of models over all subsets for (a) PathMNIST, (b) BloodMNIST, (c) OrganAMNIST, and (d) OrganCMNIST. As with the case of MNIST, correlations are generally non-trivial, with the possible exception of GNB to all other models.}
    \label{fig: medmnist heatmap}
\end{figure}

The results for the MedMNIST datasets were consistent with the results for MNIST above, with only minor differences. Most importantly, we continued to see positive correlations among the error \textit{Z}-scores of models trained on subsets of all four of these much more complex parent MedMNIST datasets---BloodMNIST, PathMNIST, OrganAMNIST, and OrganCMNIST (Fig. \ref{fig: medmnist heatmap}). The results on these datasets therefore add further support to the X-factor being an intrinsic dataset property. Finally, we tested whether the cross-architecture correlations noted above survive at the per-class level, and found almost universally high correlations for each class, e.g., for each of the digits 0-9 in MNIST, by Spearman's $\rho$ (Fig. \ref{fig: classes}), though with the details varying by digit. Further investigation is left for future work.

\section{Discussion}\label{sec: discussion}

Dataset generation and curation require significant time and energy, especially in technical and specialist areas such as medicine. Our results suggest that if a dataset results in especially high performance on a model, it is more likely because of the dataset's intrinsic features than because of some special property of the model. The most important practical implication is that as new ML models are released, they are also likely to perform especially well on such a dataset, obviating the need to review or potentially create new datasets every time a new model or architecture is introduced, potentially saving substantial effort. 

This X-factor can be understood to contribute to dataset quality. While the idea that datasets can be high or low quality is not new, the standard correlates of quality have been size and class balance (with procedural contributors to quality such as correct labeling or being in focus/in frame a given). In contrast, here we demonstrate a model- and architecture-independent factor beyond those properties that predicts performance, and hence also contributes to quality. A 2024 study \citep{couch2024} proposed sentropic measures as a way to quantify dataset quality but did not control for size and class balance, leaving open the question of how independent these (or other) factors might be as contributors to quality. Here we explicitly control for size and class balance and, by testing a variety of models, we provide strong evidence that the additional variance in model performance (which is large under our selection strategies as selection strength is adjusted) is determined by intrinsic, model-independent features---what we call the X-factor. In this way, this work provides a targeted empirical foundation for efforts trying to identify universal targets for data optimization.

Falsification required comparing selected (i.e., non-random) subsets; uniformly random subsets fail to falsify our null hypothesis by themselves (\cref{fig: subsets}c). The explanation is that for random subsets, any characteristic or feature of the dataset, including those that influence the error independent of size and class balance, becomes a random variable under any sub-sampling regime. These will concentrate strongly at their mean at any non-trivial size due to the law of large numbers. Selected subsets address this by moving the sampling mean directly rather than probing random subsets' (very small) variance. For random subsets, model architecture and training-set size explain nearly all the variance in performance as a whole, through patterns consistent with known scaling laws. Random subsets seem special in this respect; moreover, we demonstrated that the likelihood that they perform incrementally better or worse than expected seems to follow the same non-normal scaled distribution, the origin of which remains unexplored. Model-to-model stochasticity is a possibility for some models but not others; in any event, the extensive training procedure for the neural nets is expected to have minimized such variation in models that might be expected to be prone to it (deep neural networks).

The issue of residual variation also arises in Fig. \ref{fig: subsets}a-b, which demonstrated that correlations in performance between models, while high (mean $R^2$ 0.82), were not perfect. That figure also showed two blocks of substructure, in which the MLPs (perhaps not surprisingly) correlated very highly with each other in one block and the ResNets (more surprisingly) correlated very highly with RF and SVC in the other. Interestingly, this pattern was not obvious in the correlations in performance for the individual classes; there, if anything, RF and SVC correlated more strongly with the MLPs. While we largely ignore GNB as the non-expressive exception that proves the rule for most of our observations, we do note it correlates fairly well with most of the other models by class.

We demonstrated the robustness of the X-factor by testing many different datasets, and showed what was true for the simplest---MNIST---was true for the others. Among the few differences between the MNIST and MedMNIST datasets, we observed that the relative performance of the models varied slightly for different parent datasets (see Appendix (\ref{sec: medmnist_res})). This is perhaps unsurprising given the highly diverse set of contexts and modalities they represent. It could indicate that certain models are better suited for some tasks (e.g., classifying pathology slides) and worse for others (e.g., classifying handwritten digits). Note that this is independent of the conclusion that dataset quality is model-independent. We also noted that ViTs occasionally yielded uncharacteristically low-performing outliers (e.g., the pink dots at 30 images per class in Fig. \ref{fig: performance}a). However, training new ViTs on these outliers restored more typical performance, suggesting this was some form of non-reproducible random training failure idiosyncratic to this model, unrelated to intrinsic dataset quality.

It is useful to note how our results compare to work on data pruning and coreset selection. The goal of these areas is to develop algorithms that select subsets of the training data that maximize performance for their size. The underlying assumption is that not all datapoints contribute equally to performance: some are more informative---for example datapoints at decision boundaries, ones that lead to high coverage in feature space, or datapoints whose gradients match label gradients---while others are redundant or otherwise uninformative \citep{Sorscher:2022, griffin:2024, killamsetty:2021}. While our goal was not to maximize performance but simply to increase variability in performance, our selection strategies also resulted in some subsets with better-than-random performance. Some of these strategies are reminiscent of ideas explored in the coreset literature. What is new is the demonstration that the variance introduced by different selection strategies has the same effect across very different model architectures, regardless of whether selection resulted in better subsets (as they would in coreset selection) or ones that perform worse. Moreover, we identify a commonality to the effect that to our knowledge has not been reported previously and will be interesting to investigate for future work (Fig. \ref{fig: distr}).

\subsection{Limitations and Future Work}\label{sec: limitations}

The sources and explanation for the correlations falling short of $R^2=1$ and for the variation in the patterns seen by class remain unanswered and suggest potentially interesting avenues for future work, although run-to-run variance in the non-deterministic models could play at least some role. 

 Our findings could potentially be strengthened by including additional subset-selection strategies. Note the strategies included were designed to move the mean error (of any given model) in order to circumvent the law-of-large-numbers limitation to random subsets (see \cref{sec: results}), not to test any particular hypothesis about the nature of the X-factor nor to optimize performance. It might be useful to explicitly include strategies designed to produce better performance \citep{Sorscher:2022}, although this was beyond the present scope. Similarly, it could be valuable to test additional datasets beyond MNIST and MedMNIST, which were included as a standard example and for medical applications, respectively, and to explore tasks other than classification. 
 
 Expanding this study to include larger training sets (by considering, for example, ImageNet as the source dataset) is conceptually straightforward; however, it may not be computationally feasible with the current methodology. In particular, the ``uniqueness'' selection strategy does not scale well with size. It is similarly straightforward, given a similarity metric between images, to compute sentropic \citep{nguyen2026, Friedman:2022} or other measures to determine whether or how well any of them might account for the correlations observed here. Both of these are left for future work.

 A logical next step is to go beyond demonstrating the existence of this X-factor as a dataset-intrinsic property and defining what it is. For example, it would be interesting to measure these subsets sentropically as proposed by \citep{couch2024} and/or to seek theoretical results. As a result, we cannot yet say what happens to quality when, for example, two high-quality subsets are combined: whether X-factors are additive vs. super-additive or whether they potentially even interfere with each other, resulting in lower quality in the combined set. And while some strategies fortuitously improve on baseline dataset performance, directed investigation of how to optimize dataset quality by tuning the X-factor is also left for future work.

 We also note that quality can affect criteria other than accuracy, sometimes in conflicting ways. Examples of these other criteria include calibration, robustness, subgroup performance, out-of-distribution generalization, and fairness. Here we have focused on accuracy as a universal and well-defined criterion of great and clear importance for machine learning. Other criteria are left for future investigation.

\section{Conclusions}

Overall, this study provides strong evidence against our null hypothesis and for the existence of a dataset-intrinsic X-factor contributing to model performance, a newly identified contributor to dataset quality. Further defining and optimizing this factor is therefore expected to provide an independent mechanism for improving classifier performance, enabling a more fundamental understanding of how and what models learn, toward their optimization.

\vspace{6pt} 

\funding{This work was supported by the Gordon and Betty Moore Foundation (grant number?) and by the NIH under grants R01HL150394, R01HL150394-SI, R01AI148747, and R01AI148747-SI. }

\appendixstart
\appendix
\section{Appendix}

\subsection{Model details}

\subsubsection*{Random forest (RF)}

The random forest classifier implementation from scikit-learn (\verb|sklearn.ensemble.RandomForestClassifier|) is used with all default parameters, i.e.:
\begin{verbatim}
model = RandomForestClassifier()
model.fit(X,y)
\end{verbatim}
All scikit-learn models in this work are invoked this way.

\subsubsection*{Support-vector classifier (SVC)}

We used scikit-learn's (\verb|sklearn.svm.SVC|) with default parameters.

\subsubsection*{Gaussian naive Bayes (GNB)}

We used scikit-learn's (\verb|sklearn.naive_bayes.GaussianNB|) with default parameters.

\subsubsection*{K-Nearest Neighbor Classifier (KNN)}

We used scikit-learn's (\verb|sklearn.neighbors.KNeighborsClassifier|) with three different \verb|n_neighbors|: 5, 10, and 30. All other parameters were default.

\subsubsection*{Multilayer perceptrons (MLPs)}

Using torch, we create MLPs of width \verb|width| and depth \verb|depth| with a single all-to-all input layer taking the \verb|input_dim=|$28\times 28$ input features and mapping them to a \verb|width| dimensional space. This is followed by a ReLU activation function and \verb|depth-1| additional all-to-all linear layers taking an input and output dimension of \verb|width|, each followed by a ReLU. We then append a final all-to-all layer with input dimension \verb|width| and output dimension 10 (the number of classes). We will refer to the depth-1 through depth-5 MLPs as MLP-1 through MLP-5.

\subsubsection*{Resnet}

We start with the pre-trained resnets \cite{resnet18} available from torchvision \cite{torchvision} as \verb|torchvision.models.resnet18| and \verb|torchvision.models.resnet34|, both with default weights. We then overwrite the first (with \verb|torch.nn.Conv2d|) and last (with \verb|torch.nn.Linear|) layers of these models to correct for MNIST images being greyscale and having only 10 classes.

\subsubsection*{VisionTransformer (ViT)}

We created our model from a deep learning library PyTorch Image Models (\verb|timm|) \cite{rw2019timm}. To have training finish in a reasonable timeframe, we used the  \verb|vit_tiny_patch16_224.augreg_in21k_ft_in1k| model, which is tiny vit (\cite{wu2022tinyvitfastpretrainingdistillation}) pretrained on on ImageNet-22k with distillation and fine-tuned on ImageNet-1k. The weights were originally based on 224 x 224 image size, so we leveraged the \verb|dynamic_img_size| and \verb|dynamic_img_pad| features in \verb|timm| to accommodate for our parent datasets, which are all 28 x 28. Datasets' original number of classes and color channels were also preserved.

\subsection{Selection strategies}\label{app: selection}

\subsubsection*{$a$: Uniqueness}

Positive selection eliminates images from each class that seem redundant (selection is for uniqueness). Negative selection keeps only the most redundant images.

For each class, beginning with $n+m$ images, positive selection first identifies the pair of images that has the smallest RMSD. For each image in this pair, the distance to its second nearest neighbor is computed. The image with the smallest distance is eliminated from the pair. This process repeats until only $n$ images remain. 

In negative selection, first the pair of images with the \textit{largest} RMSD is identified. For each image in this pair, the distance to its second \textit{furthest} neighbor is computed and the image with the \textit{largest} distance is eliminated from the pair. Again, the process repeats, until only $n$ images remain.



\subsubsection*{$b$: Global mean}

For each class, this function first calculates the pixelwise mean image over all 6,000 images of that class in the MNIST training set, resulting in 10 mean images, one for each digit (i.e. one for each class). Then, for each of the subset's $m+n$ images, we compute the RMSD between it and each of the 10 mean images. We then average these ten RMSD distances and associate this distance to the image. In positive selection, we keep the $n$ images with the largest RMSD, maximizing distance from the global mean. In negative selection, we keep the $n$ images with the smallest RMSD.

\subsubsection*{$c$: Other mean}

This function starts with the same mean images as in $b$. For each class, for each image $i$ in that class, we compute the RMSD to the mean image for each of the \textit{other} classes. We take the maximum of these RMSDs and associate this value with image $i$. In positive selection, for each class we keep the $n$ images with the highest values, the idea being to keep the images that in a sense ``look least like'' other digits. In negative selection, we keep the $n$ images with the lowest values, i.e. those that look most like other digits.

\subsubsection*{$d$ Own mean}

This function also starts with the mean images in $b$. For each class, for each image in that class, we compute the RMSD to the mean image for \textit{its own} class. In positive selection, we keep the $n$ images with the smallest RMSD, i.e. the ones most like their own class mean; in negative selection, we keep the images with the largest RMSD, i.e. the ones least like their own class mean.


\subsection{Hardware}\label{sec: hardware}

All experiments were run on an Ubuntu workstation with an AMD Ryzen Threadripper PRO 7995WX 96-Cores CPU, 8 Samsung DIMM Synchronous Registered (Buffered) 5600 MHz (0.2 ns) memory cards, and two NVIDIA RTX 6000 Ada Generation graphics cards. 

In the example of MNIST, generating the 500 uniformly random subsets took less than a second, and generating the 2000 selected subsets took around 161 minutes. Training the four classical models from sklearn took about an additional 97 minutes. The resnet-18 and resnet-34 models were re-trained in ~244 and ~393 minutes respectively. Training the multi-layer perceptrons took between about 75 and 180 minutes each. Finally, ViT took around 360 minutes to finish. Thus the entire pipeline took a bit longer than a day to run.

It should be noted that several variations of the current experimental pipeline were considered and run, and the results analyized, prior to settling on the current version.

\subsection{Assets and Licenses}\label{sec: assets}

\subsubsection*{MNIST dataset}
This work makes use of the MNIST handwritten digits dataset \cite{mnist}. The MNIST Dataset's copyright is owned by Yann LeCun and Corinna Cortes, which  is made available under the terms of the Creative Commons Attribution-Share Alike 3.0 license. More details in \url{http://www.pymvpa.org/datadb/mnist.html}. For the convenience of coding, we imported this dataset from PyTorch's torchvision (license in a later section)

\subsubsection*{Models from sklearn} 
We used models from scikit-learn \cite{scikit-learn}, including Random Forest, Support-vector classifer,  and Gaussian naive Bayes. Scikit learn package is distributed under BSD 3-Clause License. More details \url{https://github.com/scikit-learn/scikit-learn?tab=BSD-3-Clause-1-ov-file}

\subsubsection*{Assets from torchvision}
We used torchvision \cite{torchvision} for the MNIST dataset and resnet model with pretrained weights on ImageNet, all under the BSD 3-Clause License. More details at \url{https://github.com/pytorch/vision/blob/main/LICENSE}

\section{Detailed Results} \label{sec: medmnist_res}

\subsection{Error by size}

\begin{figure}[ht]
    \centering
    \includegraphics[width=0.83\linewidth]{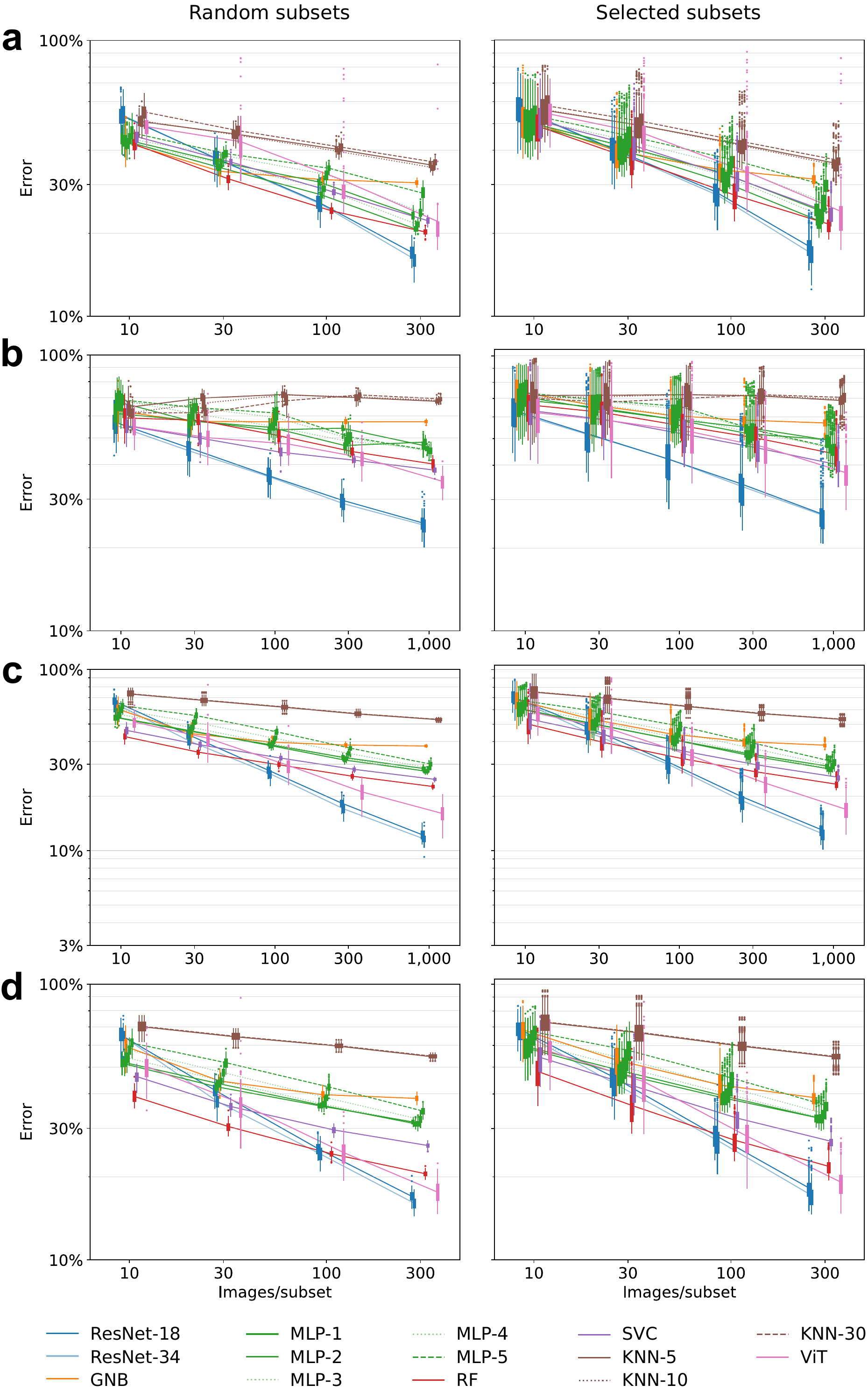}
    \caption{Performance of models on all random (left) and selected (right) subsets vs. subset size for (a) BloodMNIST, (b) PathMNIST, (c) OrganAMNIST, and (d) OrganCMNIST.}
    \label{fig: medmnist performance}
\end{figure}

\begin{figure*}[!ht]
    \centering
    \includegraphics[width=1.\linewidth]{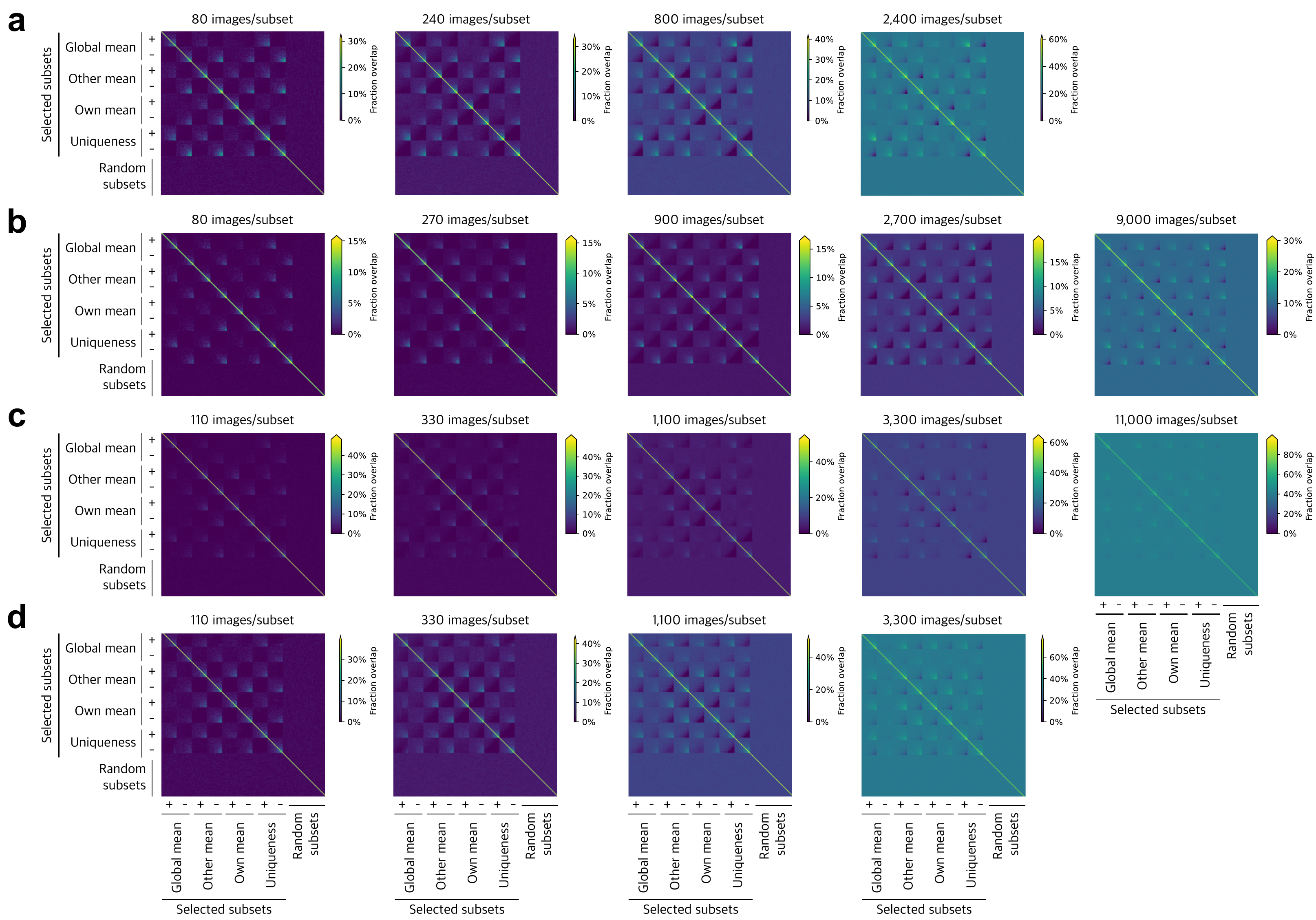}
    \caption{Percent of images that pairs of subsets have in common for subsets of the listed numbers of images per class for (a) BloodMNIST, (b) PathMNIST, (c) OrganAMNIST, and (d) OrganCMNIST. Positively selected subsets share images more than other pairs, yielding the faint checkerboard pattern. The color scale is saturated at the upper end to better visualize these effects.}
    \label{fig: medmnist overlap}
\end{figure*}

As with MNIST, we see the well known power law relationship between performance and size on the MedMNIST datasets (\cref{fig: medmnist performance}). However, the errors we see for the MedMNIST datasets are generally larger. This would seem to indicate that the classification tasks associated to these datasets are more difficult than that of classifying handwritten digits. We do see for certain parent datasets (particularly BloodMNIST) and certain subset sizes a large number of ViT models whose performance is far below the trend. We believe this is due to a random failure of the models to effectively train. To test this, we re-ran our ViT training code on the six BloodMNIST subsets with the lowest performance, and observed that all but one led to models with substantially better performance. The remaining subset also led to a more performant model after another retry.

\subsection{Error by Selection}

It remains true for the MedMNIST subsets that the selected subsets generally lead to lower performing trained models than randomly selected models, with a few exceptions. For example, for Blood, for subsets with 300 images per class (the largest size for our BloodMNIST subsets), only the 'own mean' strategy in the positive direction seems to beat random selection for training RF models, and only for selection strengths in a particular range. For PathMNIST subsets  with  1000 images/class, we find instead that the 'own mean' with negative selection produces the best selected subsets for training RF models, but once again the (average) error is not monotinic with selection strengths. For OrganAMNIST subsets of size 1000 images/class, the positive selection variant of both the uniqueness and own mean strategies seems to perform comparably, and seem to outperform the average from random selection for at least all sufficiently large strengths we tested. Thus, whatever is being tweaked by these strategies (e.g. a diversity, like one from the LCR \cite{leinster2021} framework) seems to matter differently for these different classification tasks. This does not however contradict our claim that quality is model independent (even if it does, perhaps unsuprisingly, turn out to be task dependent).

\subsection{Distribution of Z-scores}

Our observation that the performance of models trained on randomly selected MNIST subsets seem to be identically distributed, up to a shift and re-scaling accounting for the mean and standard deviation, regardless of size and model type, does seem to generalize to the MedMNIST subsets. ViT does seem at first glance to violate this trend, though we believe this is an artifact of the outliers mentioned above. 

\subsection{Subset overlap}

One concern might be that at least part of the correlation in performance between different model types trained on the same subsets might be due to overlap in the content of the subsets. However, as can be seen in the main text, the overlaps are not larger than one would expect given the finite sizes of the parent datasets.

\begin{adjustwidth}{-\extralength}{0cm}

\reftitle{References}

\bibliography{ref}

\end{adjustwidth}
\end{document}